\documentclass{llncs}
\usepackage[T1]{fontenc}
\usepackage{graphicx}
\usepackage{booktabs}
\usepackage{multirow}
\usepackage{array}
\usepackage{hyperref}
\usepackage{xcolor}
\usepackage{tcolorbox}
\usepackage{makecell}

\begin{document}

\title{Evaluating Large Language Models for Multilingual Abstractive Summarization: A Comprehensive Study of English and Bengali}
\titlerunning{Evaluating LLMs for Multilingual Summarization}

\author{
Khan Mahi Al Atahar\textsuperscript{*} \and
Nazifa Tasneem\textsuperscript{*} \and
Ajwad Abrar
}

\authorrunning{Atahar et al.}

\institute{
Department of Computer Science and Engineering, Islamic University of Technology, Gazipur, Bangladesh\\
\email{\{mahiatahar, nazifatasneem, ajwadabrar\}@iut-dhaka.edu}\\
\textsuperscript{*}These authors contributed equally to this work.
}

\maketitle              

\begin{abstract}
Abstractive summarization remains challenging across languages, where linguistic quality and factual consistency may diverge, particularly in low-resource settings. We present a standardized zero-shot benchmark of eight closed-source and open-weight LLMs on size-matched English and Bengali XL-Sum test sets. Our evaluation combines lexical and semantic metrics with reference-free factuality verification and multidimensional LLM-as-a-Judge assessment, providing a comprehensive analysis of both surface-level quality and factual reliability in multilingual summarization. Results reveal substantial cross-lingual variation: GPT-5.6 Luna leads several English lexical metrics but does not maintain the same advantage in Bengali, where LLaMA~3.3~70B achieves the highest ROUGE and BLEU scores. Gemini~2.5~Flash obtains the highest observed Bengali factuality score. Furthermore, evaluation methods produce different model rankings, highlighting the limitations of relying on a single metric for multilingual summarization assessment. These findings demonstrate that strong English summarization performance does not necessarily transfer to Bengali and emphasize the need for comprehensive evaluation frameworks for low-resource languages.


\end{abstract}

%
%
\section{Introduction}\label{sec:intro}

Abstractive news summarization aims to turn long news articles into short summaries while preserving important facts and meaning~\cite{hasan2021}. Recent advances in Transformer-based sequence-to-sequence models have substantially improved abstractive summarization. Models such as BART and PEGASUS demonstrated strong performance on abstractive text generation and summarization tasks \cite{lewis2020,zhang2020pegasus}. Modern closed-source and open-weight models have also demonstrated strong zero-shot summarization capabilities, reducing the need for task-specific training.

However, recent literature highlights important limitations in LLM-based summarization, particularly for low-resource languages. Benchmark performance remains heavily concentrated on high-resource languages, predominantly English, while languages such as Bengali may be affected by tokenization inefficiencies, weaker linguistic representation, and reduced semantic coverage~\cite{rust2021good,ahia2023,petrov2023unfairness}. At the same time, evaluating abstractive summaries remains challenging because widely used lexical metrics such as BLEU and ROUGE primarily measure surface-level n-gram overlap and may penalize valid paraphrases. More importantly, these metrics do not reliably capture factual inconsistencies or hallucinated information, which can substantially affect the reliability of generated summaries~\cite{farquhar2024}.

To address these challenges, this paper presents a cross-lingual empirical evaluation of eight closed-source and open-weight LLMs on the English and Bengali XL-Sum benchmark~\cite{hasan2021}. The study combines automatic evaluation metrics, factuality assessment, and LLM-as-a-Judge evaluation to examine summarization quality across both languages. The main contributions of this work are:
\begin{enumerate}
\item[(i)] We establish a standardized zero-shot benchmark for evaluating eight contemporary LLMs on English and Bengali abstractive summarization.

\item[(ii)] We provide a multi-dimensional evaluation framework combining automatic metrics, factuality assessment, and LLM-based qualitative evaluation.

\item[(iii)] We analyze cross-lingual differences in model performance to determine how reliably English summarization performance transfers to Bengali.

\end{enumerate}

%
\section{Related Work}\label{sec:related}

\subsection{LLMs in Summarization}

Instruction-tuned LLMs have shown strong zero-shot summarization capabilities without task-specific fine-tuning. Goyal et al.~\cite{goyal2022news} reported that human evaluators strongly preferred GPT-3-generated summaries over those produced by fine-tuned summarization models, while later work showed that summaries generated by state-of-the-art LLMs can achieve quality comparable to those written by freelance writers~\cite{zhang2024benchmarking}. However, most of this progress has been demonstrated primarily in English. Bengali summarization remains comparatively under-resourced, although multilingual datasets and benchmarks such as XL-Sum and BanglaNLG have provided important resources for developing and evaluating Bengali text generation systems~\cite{hasan2021,bhattacharjee2023}. 

\subsection{Automatic Evaluation}

Automatic summarization evaluation commonly relies on lexical overlap metrics such as ROUGE~\cite{lin2004rouge}, BLEU~\cite{papineni2002bleu}, and chrF~\cite{popovic2015chrf}, while BERTScore~\cite{zhang2019bertscore} provides a semantic similarity-based alternative. However, such metrics do not reliably capture factual consistency and may correlate only weakly with human judgments of summary quality~\cite{fabbri2021,maynez2020faithfulness}. To address this limitation, QA-based evaluation methods such as QuestEval~\cite{scialom2021questeval} and QAFactEval~\cite{fabbri2022} assess whether information in a generated summary is supported by the source document. BanglaSummEval~\cite{rafid2026banglasummeval} extends this approach to Bengali using a reference-free QG/QA framework, making it suitable for factuality evaluation in Bengali summarization.

\subsection{LLM-as-a-Judge for Evaluation}
LLM-as-a-Judge has recently emerged as a scalable approach for evaluating model-generated outputs across different tasks, including medical question answering~\cite{adib2025assessing}. It uses capable LLMs to assess responses based on predefined criteria and provides a flexible alternative to expensive human evaluation. Prior studies have shown that LLM judges can achieve strong agreement with human preferences~\cite{zheng2023judging}. However, LLM-based evaluation may suffer from self-preference, bias, and calibration issues~\cite{gu2024}, leading to proposals for more diverse judge panels~\cite{verga2024replacing}. Since most existing validation focuses on English, the reliability of LLM judges for low-resource languages such as Bengali remains less explored.

%
\section{Methodology}\label{sec:methods}

\subsection{Dataset}\label{subsec:dataset}


We utilize the XL-Sum dataset~\cite{hasan2021}, a large-scale multilingual abstractive summarization corpus derived from BBC news articles across 44 languages. The dataset provides document-summary pairs with professionally written reference summaries. For cross-lingual evaluation, we use the complete Bengali test split and randomly sample 1,012 examples from the English test partition using a fixed seed (\texttt{random\_state=42}) to match the number of Bengali test instances. This size-matching strategy ensures that both languages are evaluated on an equal number of samples.

\subsection{Models}\label{subsec:models}

We evaluate eight LLMs, consisting of four closed-source models and four open-weight models, as summarized in Table~\ref{tab:models}. The models were selected to provide a balanced comparison between closed-source and open-weight systems while covering multiple model families and providers. This selection enables us to examine whether summarization performance and cross-lingual behavior vary across different model ecosystems rather than being specific to a single provider. All models are evaluated in a zero-shot setting using standardized instruction prompts. Across all models, generation was limited to \texttt{max\_tokens=2048} with a temperature of 0.3. Models were accessed through their respective API providers listed in Table~\ref{tab:models}. Exact model identifiers and experimental configurations are available in the public GitHub repository linked in the Data and Code Availability section.

\begin{tcolorbox}[
    colback=gray!10,
    colframe=gray!40,
    coltitle=black,
    boxrule=0.5pt,
    arc=1mm,
    title=Zero-Shot Summarization Prompt,
    fonttitle=\bfseries
]

Summarize the following news article accurately and concisely. Preserve the key facts and important information. Do not include unnecessary details or filler.

\end{tcolorbox}

\begin{table}[t]
\caption{Overview of the eight evaluated LLMs and their access providers.}
\label{tab:models}
\centering
\small
\setlength{\tabcolsep}{12pt}
\begin{tabular}{lll}
\toprule
\textbf{Model} & \textbf{Access Provider} & \textbf{Type} \\
\midrule
GPT-5.6 Luna & OpenAI & Closed-source \\
GPT-4o Mini & OpenAI & Closed-source \\
Gemini 3.1 Flash Lite & Google AI Studio & Closed-source \\
Gemini 2.5 Flash & Google AI Studio & Closed-source \\
\midrule
LLaMA 3.3 70B & Groq & Open-weight \\
DeepSeek V4 Flash & OpenRouter & Open-weight \\
Qwen 3.6 27B & OpenRouter & Open-weight \\
Qwen 3.5 Flash & OpenRouter & Open-weight \\
\bottomrule
\end{tabular}
\end{table}

\subsection{Evaluation Framework}\label{subsec:evaluation_framework}

Our evaluation framework combines three complementary approaches to assess summarization quality from different perspectives: lexical and semantic similarity, factual consistency, and qualitative assessment using an LLM judge.

\subsubsection{Automatic Lexical and Semantic Metrics.}

We report ROUGE-1/2/L, BLEU, chrF, and BERTScore-F1. Bengali ROUGE uses Indic-aware Unicode tokenization, while English ROUGE uses Porter stemming. BERTScore uses \texttt{roberta-base} for English and \texttt{xlm-roberta-base} for Bengali; therefore, scores are interpreted only within each language. For consistent comparison, all reported metrics are converted to a 0--100 scale.

For selected model comparisons, statistical uncertainty is assessed using 95\% bootstrap confidence intervals, paired $t$-tests, and bootstrap significance estimates based on 5,000 paired resamples over the same test instances. The corresponding statistical results are reported in Table~\ref{tab:statistical_results}.

\subsubsection{Reference-Free Factuality Evaluation.}

We use a reference-free QG/QA factuality framework based on SummEval and BanglaSummEval~\cite{fabbri2021,rafid2026banglasummeval}. The pipeline first extracts salient entities and information from the generated summary and uses \texttt{gpt-4o} to generate corresponding factual questions. These questions are then answered using both the source document and the generated summary. Generated answers are compared using BERTScore-based semantic matching to assess whether information expressed in the generated summary is supported by the source document. The individual question-level results are aggregated into an F1 score, reported as SummEval F1 (SE-F1) for English and BanglaSummEval F1 (BSE-F1) for Bengali.

\subsubsection{LLM-as-a-Judge Evaluation.}

Following the LLM-as-a-Judge paradigm~\cite{zheng2023judging}, we use \texttt{gpt-5.6-terra} as an independent evaluator to assess generated summaries on a subset of the test data. Due to the computational cost of judge-based evaluation across eight models and two languages, we randomly selected 100 samples for each language using a fixed seed
(\texttt{random\_state=42}) to balance evaluation coverage with computational feasibility. The same selected instances were used across all eight models to ensure a consistent comparison. Each summary was evaluated on a 1--5 Likert scale across six dimensions: Relevance, Completeness, Faithfulness, Coherence, Conciseness, and Fluency.

%
%

\begin{table}[t]
\centering
\small
\caption{English and Bengali XL-Sum benchmark results. R-1, R-2, and R-L denote ROUGE-1, ROUGE-2, and ROUGE-L, respectively. BERTScore, SE, and BSE denote BERTScore-F1, SummEval F1, and BanglaSummEval F1, respectively. All metrics are reported on a 0--100 scale. A dash (\textemdash) indicates that the corresponding language-specific factuality score is not applicable. Bold values indicate the best performance within each language group.}
\label{tab:benchmark_results}
\scriptsize
\setlength{\tabcolsep}{4.5pt}
\renewcommand{\arraystretch}{1}

\begin{tabular}{
>{\centering\arraybackslash}p{0.75cm}
>{\raggedright\arraybackslash}p{2.8cm}
>{\centering\arraybackslash}p{0.7cm}
>{\centering\arraybackslash}p{0.7cm}
>{\centering\arraybackslash}p{0.7cm}
>{\centering\arraybackslash}p{0.75cm}
>{\centering\arraybackslash}p{0.75cm}
>{\centering\arraybackslash}p{1.0cm}
>{\centering\arraybackslash}p{0.75cm}
>{\centering\arraybackslash}p{0.75cm}
}
\toprule
Lang. & Model & R-1 & R-2 & R-L & BLEU & chrF & BERTScore & SE & BSE \\
\midrule

\multirow{8}{*}{\rotatebox{90}{English}}
& GPT-5.6 Luna & \textbf{24.12} & 5.87 & \textbf{15.96} & 2.06 & \textbf{31.51} & 86.75 & 58.05 & \textemdash \\
& Gemini 3.1 Flash Lite & 22.94 & \textbf{5.89} & 15.32 & \textbf{2.14} & 30.39 & \textbf{87.01} & 52.28 & \textemdash \\
& Gemini 2.5 Flash & 22.13 & 5.43 & 14.72 & 1.90 & 29.62 & 86.57 & 59.44 & \textemdash \\
& Qwen 3.5 Flash & 21.99 & 4.93 & 14.48 & 1.66 & 29.46 & 86.69 & 58.52 & \textemdash \\
& Qwen 3.6 27B & 21.47 & 4.85 & 13.99 & 1.68 & 28.96 & 86.57 & 56.58 & \textemdash \\
& DeepSeek V4 Flash & 21.46 & 5.28 & 14.13 & 1.82 & 29.36 & 86.30 & \textbf{63.45} & \textemdash \\
& GPT-4o Mini & 21.16 & 5.03 & 13.95 & 1.70 & 28.78 & 86.45 & 59.76 & \textemdash \\
& LLaMA 3.3 70B & 20.38 & 5.71 & 14.00 & 1.94 & 28.72 & 85.91 & 60.18 & \textemdash \\

\midrule

\multirow{8}{*}{\rotatebox{90}{Bengali}}
& LLaMA 3.3 70B & \textbf{16.23} & \textbf{5.29} & \textbf{12.44} & \textbf{1.90} & 33.30 & 85.92 & \textemdash & 58.28 \\
& GPT-4o Mini & 15.63 & 4.53 & 11.83 & 1.74 & \textbf{33.74} & \textbf{85.97} & \textemdash & 56.73 \\
& Gemini 2.5 Flash & 15.05 & 4.32 & 11.27 & 1.64 & 33.11 & 85.60 & \textemdash & \textbf{58.43} \\
& DeepSeek V4 Flash & 15.03 & 4.20 & 11.24 & 1.58 & 33.43 & 85.58 & \textemdash & 57.34 \\
& Qwen 3.5 Flash & 14.74 & 4.09 & 11.16 & 1.56 & 33.50 & 85.64 & \textemdash & 56.15 \\
& Gemini 3.1 Flash Lite & 14.04 & 3.88 & 10.93 & 1.42 & 32.15 & 85.63 & \textemdash & 49.73 \\
& GPT-5.6 Luna & 14.01 & 3.71 & 10.65 & 1.27 & 32.53 & 85.22 & \textemdash & 54.84 \\
& Qwen 3.6 27B & 13.82 & 3.65 & 10.34 & 1.37 & 32.40 & 85.28 & \textemdash & 57.02 \\

\bottomrule
\end{tabular}
\end{table}

\section{Results and Analysis}\label{sec:results}

\subsection{Automatic Evaluation Metrics}\label{subsec:automatic_results}

Table~\ref{tab:benchmark_results} presents the results based on lexical and semantic evaluation metrics, including ROUGE, BLEU, chrF, and BERTScore-F1, for both English and Bengali XL-Sum test sets. In English summarization, GPT-5.6 Luna achieves the highest ROUGE-1 (24.12\%) and ROUGE-L (15.96\%), while Gemini 3.1 Flash Lite obtains the highest ROUGE-2 (5.89\%), BERTScore-F1 (87.01\%), and BLEU (2.14). In Bengali, LLaMA~3.3~70B achieves the highest ROUGE-1 (16.23\%), ROUGE-2 (5.29\%), ROUGE-L (12.44\%), and BLEU (1.90), whereas GPT-4o Mini achieves the highest chrF (33.74) and BERTScore-F1 (85.97\%). These results demonstrate substantial variation in model rankings across languages, indicating that strong English summarization performance does not necessarily transfer to Bengali.

To assess whether selected differences in automatic evaluation scores are
statistically meaningful, we additionally perform paired significance testing
over the same test instances. Table~\ref{tab:statistical_results} reports
statistical comparisons for selected Bengali ROUGE-1 results, including
standard deviations, 95\% bootstrap confidence intervals, paired $t$-test
statistics, and bootstrap significance estimates based on 5,000 paired resamples.

\begin{table}[t]
\caption{Statistical comparison of selected model pairs on Bengali ROUGE-1
($N=1{,}012$). Bootstrap confidence intervals and bootstrap significance
estimates are based on 5,000 paired resamples.}
\label{tab:statistical_results}
\centering
\small
\setlength{\tabcolsep}{3pt}

\begin{tabular}{@{}lcccc@{}}
\toprule
\textbf{Comparison} &
\textbf{Mean $\pm$ SD} &
\textbf{95\% CI} &
\textbf{$t$} &
\makecell{\textbf{Paired $t$-test $p$ /}\\
\textbf{Bootstrap $p$}} \\
\midrule

LLaMA 3.3 70B &
16.23 $\pm$ 9.05 &
[15.69, 16.79] &
\multirow{2}{*}{2.713} &
\multirow{2}{*}{0.0067 / 0.0070} \\

GPT-4o Mini &
15.63 $\pm$ 8.27 &
[15.12, 16.13] &
& \\

\midrule

LLaMA 3.3 70B &
16.23 $\pm$ 9.05 &
[15.69, 16.79] &
\multirow{2}{*}{5.364} &
\multirow{2}{*}{$<0.0001$ / $<0.001$} \\

Gemini 2.5 Flash &
15.05 $\pm$ 7.86 &
[14.55, 15.52] &
& \\

\bottomrule
\end{tabular}
\end{table}

The difference between LLaMA 3.3 70B and GPT-4o Mini in Bengali ROUGE-1 is statistically significant
(paired $t$-test: $p = 0.0067$; bootstrap: $p = 0.0070$).
A larger difference is observed between LLaMA 3.3 70B and Gemini 2.5 Flash
(paired $t$-test: $p < 0.0001$; bootstrap: $p < 0.001$).

\subsection{Reference-Free Factuality Evaluation}\label{subsec:factuality_results}

The factuality scores in Table~\ref{tab:benchmark_results} report reference-free QG/QA evaluation results using SummEval for English and BanglaSummEval for Bengali. DeepSeek V4 Flash achieves the highest observed English SummEval F1 score (63.45\%). For Bengali, Gemini 2.5 Flash obtains the highest observed BanglaSummEval F1 score (58.43\%), closely followed by LLaMA~3.3~70B (58.28\%). Gemini 3.1 Flash Lite records the lowest factuality scores in both languages (English SE-F1: 52.28\%, Bengali BSE-F1: 49.73\%). These results indicate that higher lexical similarity does not necessarily correspond to stronger factual consistency.

\subsection{LLM-as-a-Judge Results}\label{subsec:judge_results}

We use \texttt{gpt-5.6-terra} as a judge to evaluate generated summaries across six dimensions: Relevance, Completeness, Faithfulness, Coherence, Conciseness, and Fluency. Each dimension is scored on a 1--5 Likert scale. Table~\ref{tab:judge_results} presents the judge-based evaluation results for English and Bengali summaries.

The evaluation reveals two key findings:
\begin{enumerate}
    \item[(i)] \textbf{Fluency versus Faithfulness:} All models achieve high Fluency (4.78--5.00) and Coherence (4.22--4.82) scores, while Faithfulness remains low (1.10--1.34). This indicates that fluent summaries may still contain factual inconsistencies.

    \item[(ii)] \textbf{English-Bengali Quality Gap:} English summaries receive numerically higher LLM-judge scores than Bengali summaries in Relevance (2.84--3.16 vs. 2.36--2.60) and Completeness (3.26--3.54 vs. 2.40--2.73). This pattern aligns with the cross-lingual differences observed in the automatic evaluation metrics.

\end{enumerate}

\begin{table}[t]
\caption{LLM-as-a-Judge (\texttt{gpt-5.6-terra}) evaluation using a 1--5 Likert scale. Rel: Relevance; Comp: Completeness; Faith: Faithfulness; Coh: Coherence; Conc: Conciseness; Flue: Fluency. Bold values indicate the best performance within each language group.}
\label{tab:judge_results}
\centering
\footnotesize
\setlength{\tabcolsep}{4pt}
\renewcommand{\arraystretch}{1.15}

\begin{tabular*}{\textwidth}{
@{\extracolsep{\fill}}
>{\centering\arraybackslash}p{0.7cm}
>{\raggedright\arraybackslash}p{3.45cm}
cccccc
@{}
}

\toprule
Lang. & Model & Rel & Comp & Faith & Coh & Conc & Flue \\
\midrule

\multirow{8}{*}{\rotatebox{90}{English}}
& GPT-5.6 Luna & \textbf{3.16} & \textbf{3.54} & 1.24 & 4.76 & \textbf{3.36} & \textbf{5.00} \\
& Gemini 3.1 Flash Lite & 3.04 & 3.48 & \textbf{1.34} & \textbf{4.82} & 3.14 & \textbf{5.00} \\
& DeepSeek V4 Flash & 2.92 & 3.38 & 1.30 & 4.78 & 3.06 & \textbf{5.00} \\
& Qwen 3.5 Flash & 2.96 & 3.26 & 1.28 & 4.70 & 3.16 & \textbf{5.00} \\
& LLaMA 3.3 70B & 3.02 & 3.36 & 1.24 & 4.78 & 2.80 & \textbf{5.00} \\
& Qwen 3.6 27B & 2.86 & 3.26 & 1.24 & 4.76 & 3.10 & \textbf{5.00} \\
& Gemini 2.5 Flash & 2.90 & 3.28 & 1.20 & 4.80 & 3.00 & \textbf{5.00} \\
& GPT-4o Mini & 2.84 & 3.26 & 1.18 & 4.74 & 2.90 & \textbf{5.00} \\

\midrule

\multirow{8}{*}{\rotatebox{90}{Bengali}}
& Gemini 3.1 Flash Lite & \textbf{2.60} & 2.66 & 1.26 & 4.46 & \textbf{3.22} & 4.94 \\
& GPT-4o Mini & 2.54 & 2.56 & \textbf{1.30} & 4.36 & 3.06 & 4.80 \\
& Gemini 2.5 Flash & 2.56 & 2.60 & 1.12 & \textbf{4.56} & 2.68 & \textbf{4.96} \\
& DeepSeek V4 Flash & 2.46 & 2.50 & 1.10 & 4.46 & 3.00 & 4.94 \\
& GPT-5.6 Luna & 2.36 & 2.70 & 1.18 & 4.44 & 2.70 & \textbf{4.96} \\
& LLaMA 3.3 70B & 2.52 & 2.42 & 1.24 & 4.26 & 2.96 & 4.78 \\
& Qwen 3.6 27B & 2.51 & \textbf{2.73} & 1.14 & 4.38 & 2.61 & 4.79 \\
& Qwen 3.5 Flash & 2.54 & 2.40 & 1.14 & 4.22 & 3.00 & 4.82 \\

\bottomrule

\end{tabular*}
\end{table}

%
%
\section{Discussion}\label{sec:discussion}

Our results reveal that multilingual summarization ability is not uniformly distributed across languages. Although current LLMs achieve strong English summarization performance, these capabilities do not consistently transfer to Bengali, demonstrating that language-specific evaluation remains essential for low-resource settings.

\textbf{Cross-Lingual Performance Differences.}
The ranking differences between English and Bengali indicate that observed
model performance varies substantially across target languages. Models that
perform best in English do not necessarily maintain the same advantage in
Bengali. This suggests that multilingual summarization quality may be influenced
not only by general model capacity but also by language representation, training
exposure, and adaptation to specific linguistic characteristics.

\textbf{Mismatch Between Surface Similarity and Factuality.}
The disagreement between lexical metrics and factuality evaluation highlights the limitations of relying only on overlap-based measures. A summary may achieve strong ROUGE or BERTScore by preserving similar wording while still omitting important entities or introducing unsupported information. Therefore, factuality-oriented evaluation provides complementary evidence that cannot be captured by conventional automatic metrics alone.

\textbf{Challenges of LLM-Based Evaluation in Low-Resource Languages.}
The LLM-as-a-Judge results show that fluent and coherent summaries do not always correspond to stronger factual reliability. In Bengali, judge-based scores and factuality verification sometimes produce different rankings, suggesting that automated evaluation methods may require careful calibration for low-resource languages. Combining semantic metrics, reference-free factuality verification, and judge-based assessment provides a more complete view of multilingual summarization quality.

\section{Conclusion}\label{sec:conclusions}

This paper presents a comprehensive zero-shot evaluation of eight closed-source and open-weight LLMs for English and Bengali abstractive news summarization on the XL-Sum benchmark. We propose a multi-dimensional evaluation framework combining automatic lexical and semantic metrics, reference-free factuality assessment, and LLM-as-a-Judge evaluation. Our findings show substantial cross-lingual variation in model performance: different models lead different evaluation metrics in English, while in Bengali, LLaMA~3.3~70B achieves the highest ROUGE and BLEU scores and Gemini~2.5~Flash obtains the highest observed factuality score. These results demonstrate that model rankings obtained from English-centric evaluation do not necessarily generalize to low-resource languages.

This study has several limitations, including evaluation in a single zero-shot setting, reliance on a single evaluator model within each automated evaluation pipeline, and restriction to English and Bengali BBC news summarization. Future work will explore repeated evaluations, broader language and domain coverage, multiple judge models, human assessment, and adaptation techniques such as parameter-efficient and domain-specific tuning to improve multilingual summarization reliability.

\section*{Data and Code Availability}\label{sec:availability}

All evaluation scripts and detailed metric results are available at: \url{https://github.com/khanmahijyoti/multilingual-llm-summarization}.

\begin{credits}

\subsubsection{\discintname}
The authors have no competing interests to declare that are relevant to the content of this article.
\end{credits}

%
%
\bibliographystyle{splncs04}
\bibliography{references}

\end{document}